\documentclass[a4paper,fleqn]{cas-dc}

\usepackage[authoryear,longnamesfirst]{natbib}
\usepackage{adjustbox}
\usepackage{siunitx}
\usepackage{blindtext}
\usepackage{mathtools}
\usepackage{subcaption}
\usepackage{graphicx}

\def\tsc#1{\csdef{#1}{\textsc{\lowercase{#1}}\xspace}}
\tsc{WGM}
\tsc{QE}

\begin{document}
\let\WriteBookmarks\relax
\def\floatpagepagefraction{1}
\def\textpagefraction{.001}

\shorttitle{Estimation of Sea State Parameters  from Ship Motion Responses Using Attention-based Neural Networks}    

\shortauthors{D. Selimović. et al.}  

\title [mode = title]{Estimation of Sea State Parameters  from Ship Motion Responses Using Attention-based Neural Networks}  




\author[1,2]{Denis Selimović}
[   type=editor,
    auid=1,
    bioid=1,
    orcid=0000-0003-2485-0416
]
\ead{dselimovic@riteh.hr}
\credit{Conceptualization, Methodology, Software, Validation, Investigation, Data Curation, Writing -- Original Draft Preparation, Visualization}

\author[1,2]{Franko Hržić}
[   type=editor,
    auid=1,
    bioid=1,
    orcid=0000-0003-2485-0416
]
\ead{fhrzic@riteh.hr}
\credit{Conceptualization, Methodology, Software, Validation, Investigation, Data Curation, Writing -- Original Draft Preparation, Visualization}


\author[1]{Jasna Prpić-Oršić}
[   type=editor,
    auid=3,
    bioid=3,
    orcid=0000-0002-5742-6067
]
\ead{jasnapo@riteh.hr}
\credit{Conceptualization, Methodology, Validation, Formal Analysis, Resources, Data Curation, Writing -- Review \& Editing, Supervision, Project Administration, Funding Acquisition}

\author[1,2]{Jonatan Lerga}
[   type=editor,
    auid=2,
    bioid=2,
    orcid=0000-0002-4058-8449
]
\ead{jlerga@riteh.hr}
\credit{Conceptualization, Methodology, Validation, Formal Analysis, Investigation, Resources, Writing -- Original Draft Preparation, Writing -- Review \& Editing, Supervision, Project Administration, Funding Acquisition}
\cormark[1]


\affiliation[1]{organization={Department of Computer Engineering, Faculty of Engineering, University of Rijeka},
addressline={Vukovarska 58}, 
postcode={51000 Rijeka}, 
country={Croatia}}

\affiliation[2]{organization={Center for Artificial Intelligence and Cybersecurity, University of Rijeka},
addressline={Radmile Matejcic 2}, 
postcode={51000 Rijeka}, 
country={Croatia}}

\cortext[cor1]{Corresponding author} 


\begin{abstract}[S U M M A R Y]
On-site estimation of sea state parameters is crucial for ship navigation systems' accuracy, stability, and efficiency. Extensive research has been conducted on model-based estimating methods utilizing only ship motion responses. Model-free approaches based on machine learning (ML) have recently gained popularity, and estimation from time-series of ship motion responses using deep learning (DL) methods has given promising results. Accordingly, in this study, we apply the novel, attention-based neural network (AT-NN) for estimating sea state parameters (wave height, zero-crossing period, and relative wave direction) from raw time-series data of ship pitch, heave, and roll motions. Despite using reduced input data, it has been successfully demonstrated that the proposed approaches by modified state-of-the-art techniques (based on convolutional neural networks (CNN) for regression, multivariate long short-term memory CNN, and sliding puzzle neural network) reduced estimation MSE by 23\% and MAE by 16\% compared to the original methods. Furthermore, the proposed technique based on AT-NN outperformed all tested methods (original and enhanced), reducing estimation MSE by up to 94\% and MAE by up to 70\%. Finally, we also proposed a novel approach for interpreting the uncertainty estimation of neural network outputs based on the Monte-Carlo dropout method to enhance the model's trustworthiness.
\end{abstract}



\maketitle

\section{Introduction}
\label{sec3}
\vspace{10pt}

Maintaining stable operational performance for ship routing is critical for safety and navigational efficiency. This represents a challenge for onboard decision support systems (DSSs), which consider the problem of obtaining data on the surrounding sea state and help make operational decisions more precise \cite{Faltinsen1990}. Since encountered waves affect almost all ship operations, real-time estimation of their characteristics is of crucial importance. Different approaches for collecting information about sea state data exist. Moored wave buoys, navigational radars, or satellite data are traditionally used. However, these methods do not represent a complete solution due to their limitations in availability, reliability, convenience and/or cost of maintenance \cite{Nielsen2017a}. For example, statistical-based methods can provide accurate wave height estimates, but wave period and wave direction estimates are relatively poor. On the other hand, expensive wave radars can produce accurate estimations of wave period and wave direction, but wave height is not always estimated correctly \cite{Selimovic2020}.

As a more efficient alternative to standard estimation methods, the concept of a "wave buoy analogy" intends to measure only available ship responses to infer sea state information. These result in less complexity, easier maintenance, and ultimately lower costs. Decades of research have been conducted on this concept. A concise overview of sea state estimation techniques based on ship motions is given in \cite{Nielsen2017a}. Optimization methods were developed and refined as computing resources improved. Ultimately, this results in several published research papers incorporating the speed-of-advanced problem and full-scale tests. The concept is introduced with the assumption of linearity between sea state and measured ship responses. With an appropriate mathematical model, two different estimation approaches can be found in the literature, one in the frequency domain and another directly in the time domain. The latter intends to address challenges that appear in frequency domain procedures. In the time domain, research focuses on methods that apply Kalman filtering \cite{Pascoal2009,Pascoal2017} or stepwise procedures \cite{Nielsen2015,Nielsen2016}. However, most of the research done in the frequency domain uses spectral analysis, with a mathematical relationship between the response spectra and the directional wave spectrum (DWS). The model description is given with complex transfer functions, i.e., response amplitude operators (RAOs). The corresponding DWS is then found by inverse mapping procedures. Procedures in the frequency domain are formulated so that the mathematical model is based either on the equivalence of spectral energy distribution \cite{Nielsen2006,Tannuri2003,Hinostroza2016} or the spectral moments \cite{Montazeri2016}.

Although the above-mentioned DWS estimation techniques have shown fine performance in many reports, their implementation may include some constraints. The concept of analogy is given with the assumption of linearity, and spectral analysis requires stable operating conditions where it should be a certain minimum period of stationarity to perform one. In addition, the accuracy and reliability of estimation largely depend on the accuracy of RAOs which may be less accurate or incomplete due to insufficient knowledge of the input conditions. Recently, sea state estimation was related to data-driven ML approaches to overcome the limits of conventional RAO-based methods. The idea is to build a parametrized ML model without knowledge about ship model description. A few studies have been published on the use of ML approaches to train the model. The proposed models, trained on time and/or frequency domain features, have shown sufficient accuracy in predicting wave characteristics. As the model is based only on ship motion data, the assumption is that it can be adapted to various types of ships. An example of a model-free approach to identify sea state levels is proposed by \cite{Tu2018}. The authors apply available dynamic positioning (DP) sensors to collect time-series data in four degrees of freedom (DOF). The estimation procedure includes multi-layer classifiers, and time and frequency domain features are extracted from the data with several preprocessing techniques.

As the classification in feature-based approaches is dependent on the quality and quantity of hand-engineered features, research suggested the application of DL  methods. The DL-based model for classifying the sea state levels proposed by \cite{Cheng2019} uses only raw time-series data. Here, the classification accuracy is improved using a multi-component model. Namely, convolutional neural networks (CNNs) and fast Fourier transform help extract time and frequency characteristics, while recurrent neural networks search for long dependencies in motion data. \cite{Cheng2020} expanded this work using densely connected CNN to take into account the estimation of wave direction characteristics. Additionally, to improve the interpretability and achieve even better accuracy, \cite{Cheng2020a} proposed a model based on 2D spectrogram images, which carry both time and frequency information.

The before-mentioned papers are related to DP application without ship forward-speed. On the other hand, \cite{Duz2019} and \cite{Mak2019} successfully predicted wave height and wave direction under forward-speed conditions. Ship motions in 6-DOF are used as input, and the problem was treated as a multivariate time-series regression. Results are given for several NN structures with appropriate performance comparisons. Those works are extended to identify real-time DWS using deep convolutional encoding-decoding NN in \cite{Scholcz2020}. However, the approach is focused on training models on simulated data, and then these models are used for measured data directly or applied using transfer learning methods. In \cite{Mittendorf2022} several proposed DL architectures also prove sufficient results in the case of an advanced ship in beam and following seas with no significant effect on the prediction accuracy when a different forward-speed is applied.

In this paper, we propose several DL models for sea state parameter estimation. The main focus was on improving the currently proposed models and implementing the new DL model based on AT-NN. The input data was 3-DOF ship motion time-series, and the output indicated the wave period, relative wave direction and significant wave height. As in \cite{Duz2019}, the problem was characterized as a multivariate time-series regression. The algorithms are trained with numerically simulated data generated according to the recommended guidelines. Proposed methods have been compared to state-of-the-art reports to evaluate the novelty of the research and its potential real-life applications. 

The rest of the paper is structured as follows. In Section \ref{sec2}, the data generation process is given with described simulation procedure. Next, the applied NN architectures and DL models are described. The performances of applied models were evaluated and described in Section \ref{sec3}. Conclusion remarks on the limitations of the present work and possible future works are provided in Section \ref{sec4}.

\section{Methods}
\label{sec2}
\vspace{10pt}

\begin{figure}
	\centering
	\includegraphics[clip, width=0.5\textwidth, trim=0cm 0cm 0cm 0cm]{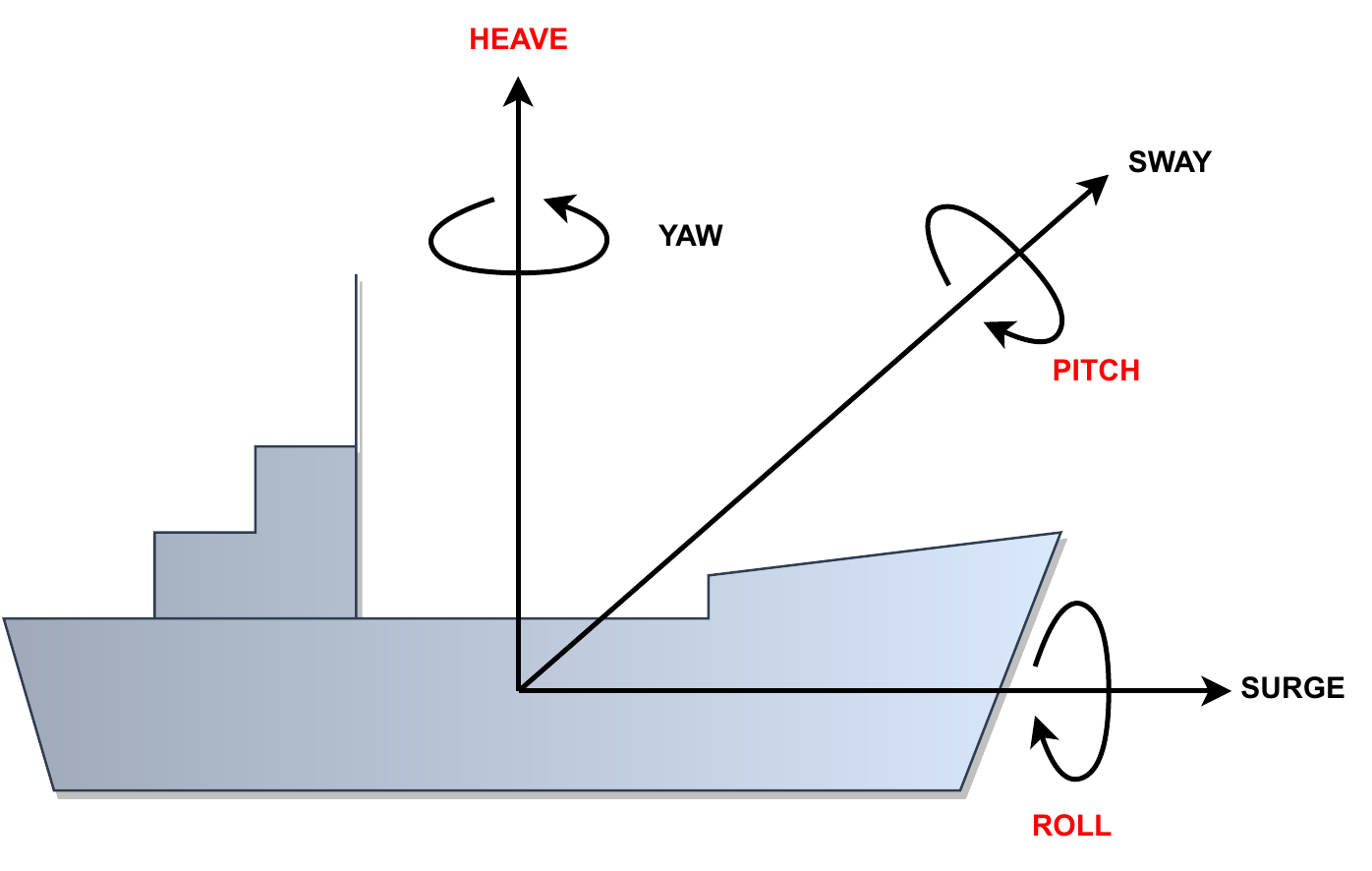}
    \caption{Ship motion responses in six degrees of freedom (6-DOF).}
    \label{fig:ShipMotions}
\end{figure}

Collecting real-time estimation of encountered wave characteristics may be worthwhile for improving navigational efficiency. While routing, the ship encounters stochastic environmental forces, thus subjected to three rotational (rolling, pitching, and yawing) and three linear (surging, swaying, and heaving) motion responses, as shown in Figure \ref{fig:ShipMotions}. According to many previous studies mentioned in the introduction of the paper, considering a set of three motion responses is advantageous for successful sea state estimation in a wave buoy analogy approach. Based on the given input-output data pairs, the idea is to create a DL model to estimate the characteristics of encounter waves with given ship motion responses in 3-DOF. We observe the roll, pitch, and heave motions, mostly driven by the sea waves and least affected by the operation of the thrusters \cite{Selimovic2020,Nielsen2006}.

For testing DL model accuracy, stationary stochastic time-series of 3-DOF ship motion responses are computed from the known wave spectra. Time-series simulation procedures are done according to recommended guidelines given in \cite{DNVGL2018}, and related research work by \cite{Mittendorf2022}. This section describes the used procedures for generating time-series data and models for wave characteristics estimation.

\subsection{Computing of time-series data}
\label{sec2_1}
\vspace{10pt}

Let us suppose the ship is navigating in the natural seaway, i.e. in irregular waves, at a specified forward speed, $U$, and with a relative heading to the waves, $\beta$. In this case wave elevation can be expressed as a sum of multiple sinusoidal (regular) wave components:

\begin{equation}\label{eq:waveElevation}
\zeta(x,t) = \sum_{i=1}^{n} A_i \cos(\omega_it-k_ix+\epsilon_i)
\end{equation}

\noindent where $\omega_i$ is a wave frequency of a particular wave component, $k_i$ is a wave number, and $\epsilon_i$ is a the random phase shift in the range $[0,2\pi]$ to ensure random nature of the waves. The amplitude $A_i$ can be found from the relation to the wave spectrum $S(\omega)$, given with equation \ref{eq:waveElevationAmplitude}:

\begin{equation}\label{eq:waveElevationAmplitude}
\frac{1}{2}A_i^2 = S(\omega_i)\Delta\omega_i
\end{equation}

\noindent where considered frequency interval $\Delta\omega_i$ has randomly distributed boundaries.

Wave spectrum $S(\omega)$ is given in the frequency domain and describes how the total wave energy of all wave frequencies $\omega_i$ is distributed in stationary seaway \cite{Faltinsen1990}. It can be estimated by one of the well-known equations suggested by the International Towing Tank Conference \cite{ITTC1969}. In the case of fully developed seas, the recommendation is to use the two-parameter Bretschneider spectrum, cf. Eq. \ref{eq:BretschneiderSpectrum}.

\begin{equation}\label{eq:BretschneiderSpectrum}
S(\omega) = 124 ~ \frac{H_s^2}{T_z^4} ~ \omega^{-5} ~ exp\left(-\frac{496}{T_z^4\omega^4}\right)
\end{equation}

\noindent The equation \ref{eq:BretschneiderSpectrum} is given in a rewritten form as the characteristic statistical period of interest is zero up-crossing period $T_z$, with a relation to mean wave period $T_1$, $T_1 = 1.087~T_z$ and peak period $T_p$, $T_p = 1.405~T_z$. $H_s$ represents significant wave height, and $\omega$ is the wave frequency \cite{DNVGL2018}. 

\begin{figure}
	\centering
	\includegraphics[clip, width=0.5\textwidth, trim=4cm 8cm 4cm 8cm]{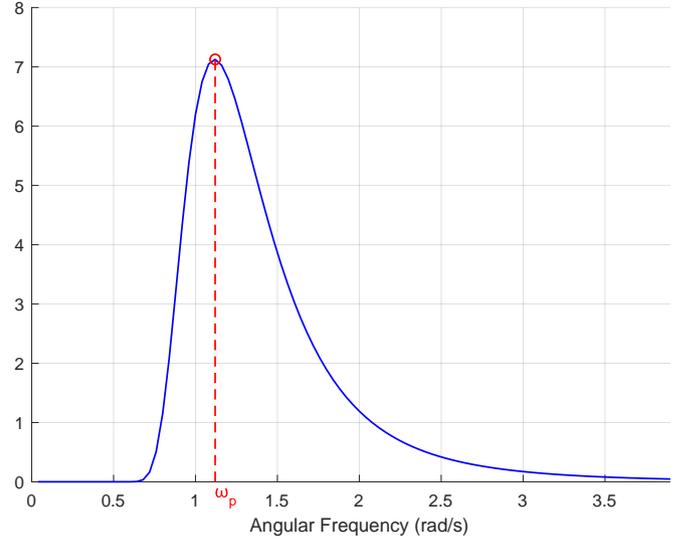}
    \caption{Unidirectional one-dimensional Bretschneider wave spectrum with $H_s=3.0~m$ and $T_z=4.0~s$.}
    \label{fig:WaveSpectraExample}
\end{figure}

The energy distribution, in an example of two-parameter Bretschneider spectrum, is given in Figure \ref{fig:WaveSpectraExample} with values of $H_s=3.0$ m and $T_z=4.0$ s. The one-dimensional spectrum has a single peak value indicating that unimodal seas are being observed. The single peak value occurs at an angular frequency $\omega_p=1.12$ rad/s, corresponding to the above relations for the peak period $T_p$ and zero up-crossing period $T_z$ since $\omega_p=2 \pi / T_p$. Also, the supplied spectrum includes information on the significant wave height $H_s$, which can be determined from the area under the spectrum. It should be noted that, for practicality and simplicity reasons, generated time-series do not contain information about wave directionality. In other words, we only observe long-crested or unidirectional waves with a narrow spreading.

While routing, the ship is observed as a complex object with the forward speed $U$, and it is encountered by waves at a specific encounter angle $\beta$. Encountered wave spectrum is not equal to wave spectrum given in wave frequency domain $\omega$ due to Doppler shift. Therefore, ship responses should be observed in the encounter frequency domain $\omega_e$. The following expression applies to the transformation from wave to encounter frequency domain:

\begin{equation}\label{eq:encounterFrequency}
\omega_e = \omega - \frac{\omega^2~U}{g} ~ \cos(\beta)
\end{equation}

\noindent with $g$ being the gravitational constant in the adopted right-handed coordinate system, where the encounter angle of $180 deg$ indicates head seas. 

With known encounter wave spectrum and appropriate complex-valued transfer functions $\Phi_{R}$, one can calculate the time-series of ship motion responses $R_i(t)$ following \cite{Denis1953}:

\begin{equation}\label{eq:motionSeries}
R_i(t) = \sum_{i=1}^{n} A_i |\Phi_{R}(\omega_{e,i})| \cos(\omega_{e,i}~ t + \epsilon_i)
\end{equation}

\noindent where frequency discretisation is non-equidistant, and amplitude $A_i$ is given in wave frequency domain $\omega$ as stated in Eq. \ref{eq:waveElevationAmplitude}. Since the unidirectional spectrum is observed, each set of time-series $R_i(t)$ is generated for a single encounter angle $\beta$, with other wave characteristics contained in the amplitude $A_i$.

\begin{figure}
	\centering
	\includegraphics[clip, width=0.5\textwidth, trim=4cm 8cm 4cm 8cm]{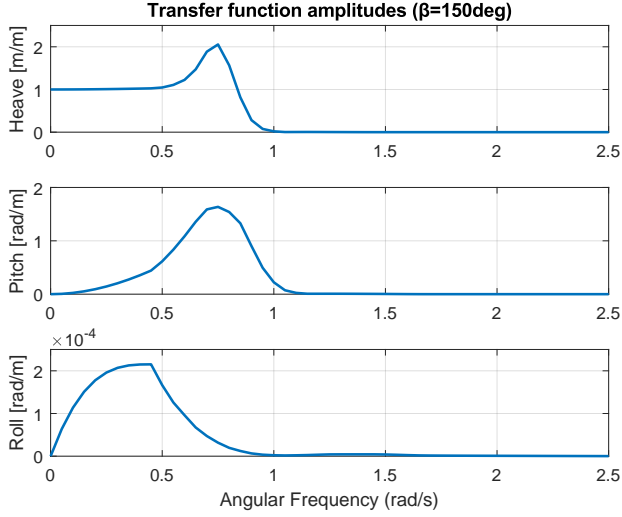}
    \caption{Amplitudes of transfer functions for heave, pitch and roll responses with relative wave direction $\beta=150$~deg. and ship forward speed $U=16.19$~knots.}
    \label{fig:RAOExample}
\end{figure}

Transfer functions are computed with a strip theory solver for pitch, roll and heave as a function of encounter frequency $\omega_e$ and encounter angle $\beta$. Forward speed was $16.19$ knots, and relative wave directions were discretized into 36 headings for $[0, 360]$deg, with other values obtained by linear interpolation. The range of wave frequencies $\omega$ are defined as $[0, 4.0]$rad/s. The ship under study was a 20,000DWT Bulk Carrier, with the primary characteristics shown in Table \ref{table:ShipParticulars}. The amplitudes of frequency response functions for heave, pitch, and roll motions are given in Figure \ref{fig:RAOExample} with relative wave direction $\beta=150$ deg and ship forward speed $U=16.19$ knots.

\begin{table}
    \caption{Main characteristics of the Bulk Carrier ship.}
    \label{table:ShipParticulars}
    \resizebox{\columnwidth}{!}{
    \begin{tabular*}{\tblwidth}{@{} c|LLL@{}}
   \toprule \toprule
    \textbf{Name}	& \textbf{Value}\\
    \midrule
    Length $L_{PP}$		& 160.4 m \\
    Breadth $B$			& 27.2 m \\
    Draugth $T$			& 6.55 m \\
    Metacentric height $G_M$			& 3.97 m \\
    Block coefficient $C_B$			& 0.77 \\
    Displacement			& 22,193 t \\
   \bottomrule \bottomrule
  \end{tabular*}}
\end{table}

\begin{figure}
	\centering
	\includegraphics[clip, width=0.5\textwidth, trim=4cm 8cm 4cm 8cm]{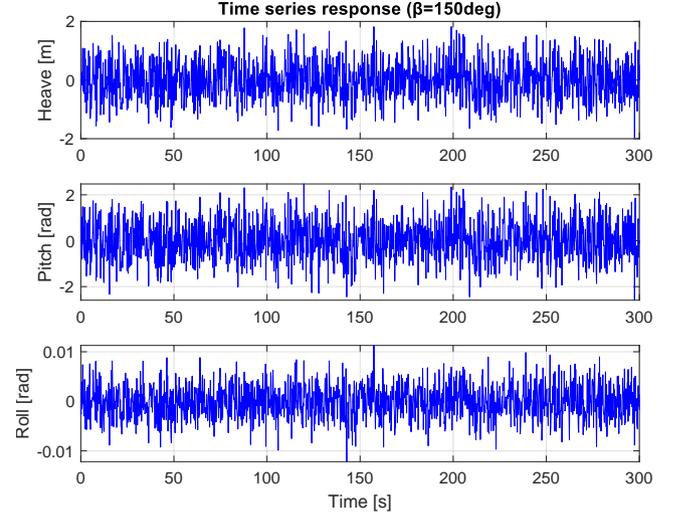}
    \caption{Calculated time-series of ship motion responses at ship forward speed $U=16.19$ knots and sea state parameters $H_s=3.0$ m, $T_z=4.0$ m and $\beta=150$ deg.}
    \label{fig:MotionsExample}
\end{figure}

By applying the Eq. \ref{eq:motionSeries}, we computed time-series of motion responses in 3-DOF, with a sampling frequency of $5$ Hz and sample lengths of $300$ s. An example of time-series responses is given in Figure \ref{fig:MotionsExample}. We design Latin hypercube sampling (LHS) to discretize the three-dimensional search space of $H_s$, $T_z$ and $\beta$ with a $20,000$ number of sample points at a uniform distance. The range of validity for parameter $H_s$ is $[0.5,10.5]$m and $[3.5,9.6]$m for parameter $T_z$. The encounter angle $\beta$ is defined in range $[0, ~360]$ deg. In addition, following \cite{Mittendorf2022} work, constraints for irregular waves are applied to include wave breaking conditions in the sense of moderate wave steepness with a limited maximum value $S_s$ given as \cite{DNVGL2018}:

\begin{equation}\label{eq:waveStepness}
S_s = \begin{cases}
\frac{1}{10} &\text{for $T_z\leq6~ s$}\\
\frac{1}{15} &\text{for $T_z\geq12~ s$}
\end{cases}
\end{equation}

\begin{figure}
	\centering
	\includegraphics[clip, width=0.5\textwidth, trim=3cm 8cm 3cm 8cm]{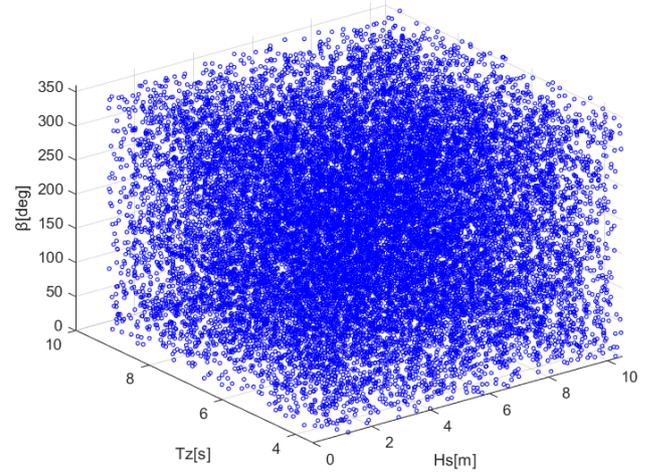}
    \caption{Wave spectra parameters $H_s$, $T_z$ and $\beta$ uniformly distributed using Latin Hypercube Sampler.}
    \label{fig:LHSExample}
\end{figure}

Finding the values between the limits given in the above equation is done with linear interpolation. The dataset contained about $13,000$ samples after these constraints were applied. The final three dimensional search space of $H_s$, $T_z$ and $\beta$ designed by LHS is presented in Figure \ref{fig:LHSExample}. The following is a detailed description of the data augmentation process and the NN architectures which are used to create the DL models.

\begin{figure*}
	\centering
	\includegraphics[clip, width=\textwidth, trim=0cm 0cm 0cm 0cm]{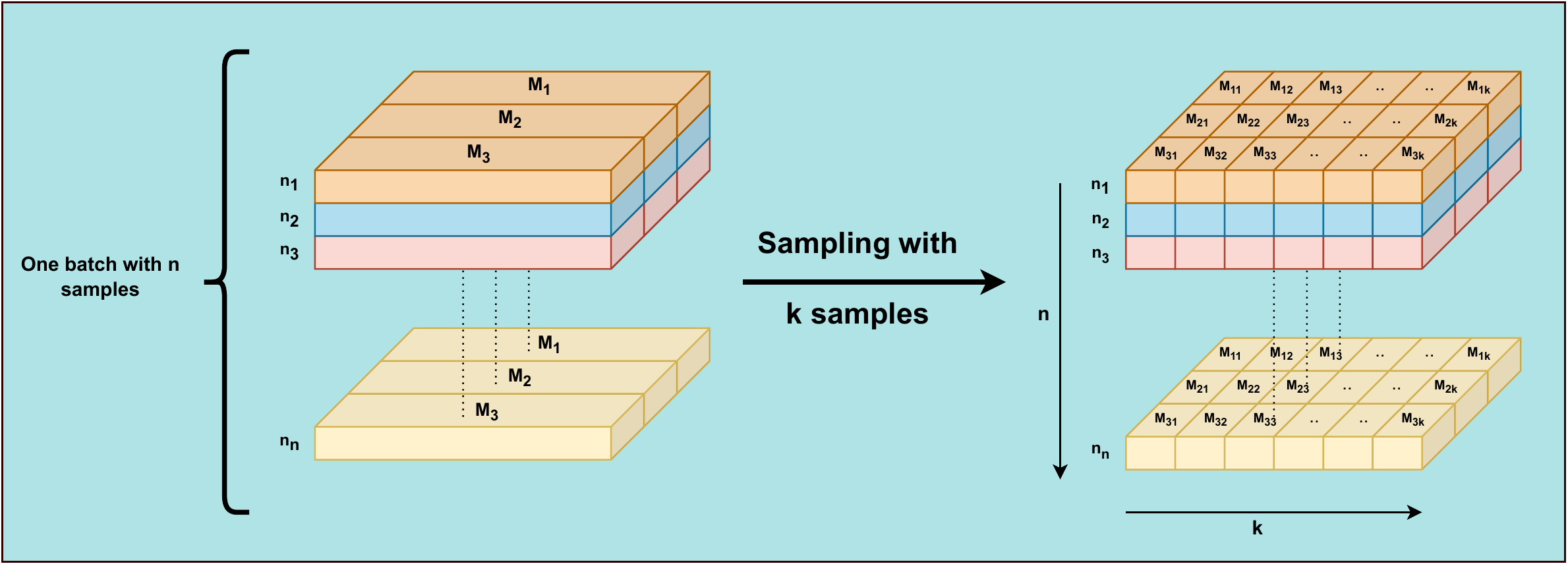}
    \caption{Data augmentation modelling with more samples.}
    \label{fig:DataAugumentation}
\end{figure*}

\begin{figure}
	\centering
	\includegraphics[clip, width=0.48\textwidth, trim=0cm 0cm 0cm 0cm]{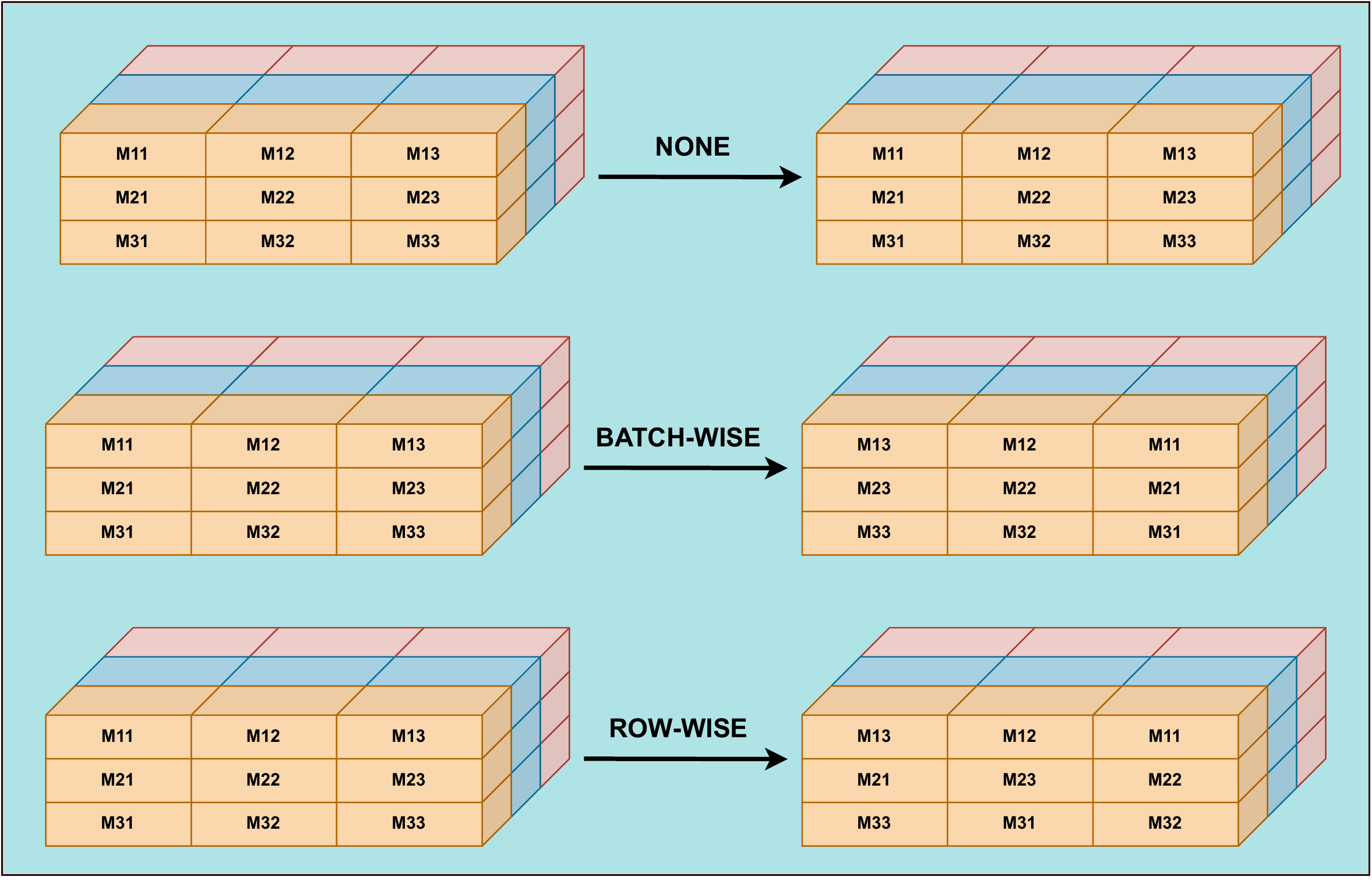}
    \caption{Comparison of batch-wise and row-wise data augmentation.}
    \label{fig:DataAugumentationDiff}
\end{figure}

\subsection{Data preprocessing}
\label{sec2_2}
\vspace{10pt}

Input-output samples were prepared for models' training in the form of JSON files, and subsequently processed and scaled. The output values of the sea state parameters were scaled so that $H_s$ and $T_z$ were divided by a value of $15$, and the relative wave direction $\beta$ values were divided by the value of $360$. The extensive experimentation showed that the scaling of input data (3-DOF) is not influencing the convergence of the tested models. However, the convergence of models is greatly influenced by the output scale (set to be between 0 and 1). By choosing scaling values of $15$ and $360$ as maximum values in our dataset enlarged by $10\%$, we have ensured the fulfillment of these conditions.  

Data augmentation strategies are employed in the data preprocessing step to increase the accuracy of our models. With data augmentation applied to proposed models, we increase the variety of our data and eliminate overfitting. 

Next, we explain the utilized data augmentation methods. Figure \ref{fig:DataAugumentation} shows one batch with $n=32$ samples. Each sample in the batch consists of three time-series input data of ship motion responses, denoted as $M_1$, $M_2$ and $M_3$. Within each batch, sampling with size $s = 32$ was performed resulting in $k = 47$ sub-samples of each input data. 

Figure \ref{fig:DataAugumentationDiff} compares transformations on how the sub-samples of each input data are shuffled. We distinguish two different approaches: batch-wise and row-wise types of augmentation. Comparison is given using one batch with $n=3$ samples and $k=3$ tokens. As depicted in the figure, batch-wise augmentation randomly shuffles the input data signals so that all rows shuffle simultaneously in the same manner. Row-wise augmentation, on the other hand, shuffles each data sample separately. To compare the results, all DL models were trained without any augmentation, as well as with batch-wise and row-wise data augmentation. The trained models always outperformed those not using the above data augmentations. 

\subsection{Deep learning models}
\label{sec2_3}
\vspace{10pt}

In this research, we propose a novel DL model for sea state parameter estimation based on the current state-of-the-art module called Attention~\cite{NIPS2017:SelfAttention}. To prove that the proposed model is fine-tuned and can be treated as the best currently available solution, we have compared it with the three other DL models presented in ~\cite{Mak2019}. Furthermore, to correctly compare the proposed model with existing models, we have enhanced the existing models by scaling the number of kernels and neurons. Here we find the work done by \cite{Mak2019} to be the most relevant due to several facts: their approach is well defined and enables their models' reproductivity, proposed models are based on different modules, and last but not least, they have been dealing with the same problem as the one addressed in this paper. The only difference between our utilization of existing models and their original usage was the number of input signals. Namely, in the original paper, the existing models utilized 6-DOF of ship motion, while our experiments only used 3-DOF. By reducing the number of DOF, we did not impair the model's integrity in any way, which is supported by the obtained results which are comparable to the results in the original paper.

In the following subsections, the utilized DL models will be presented and described in detail, while the source code implementing the models can be found in the following repository \cite{Github2022}. Furthermore, at the end of this subsection, we provide a short summary of all models' hyperparameters. 

\subsubsection{CNN for regression: original and enhanced}
\label{sec2_3_1}
\vspace{10pt}

\begin{figure}
	\centering
	\includegraphics[clip, width=0.5\textwidth, trim=0cm 0cm 0cm 0cm]{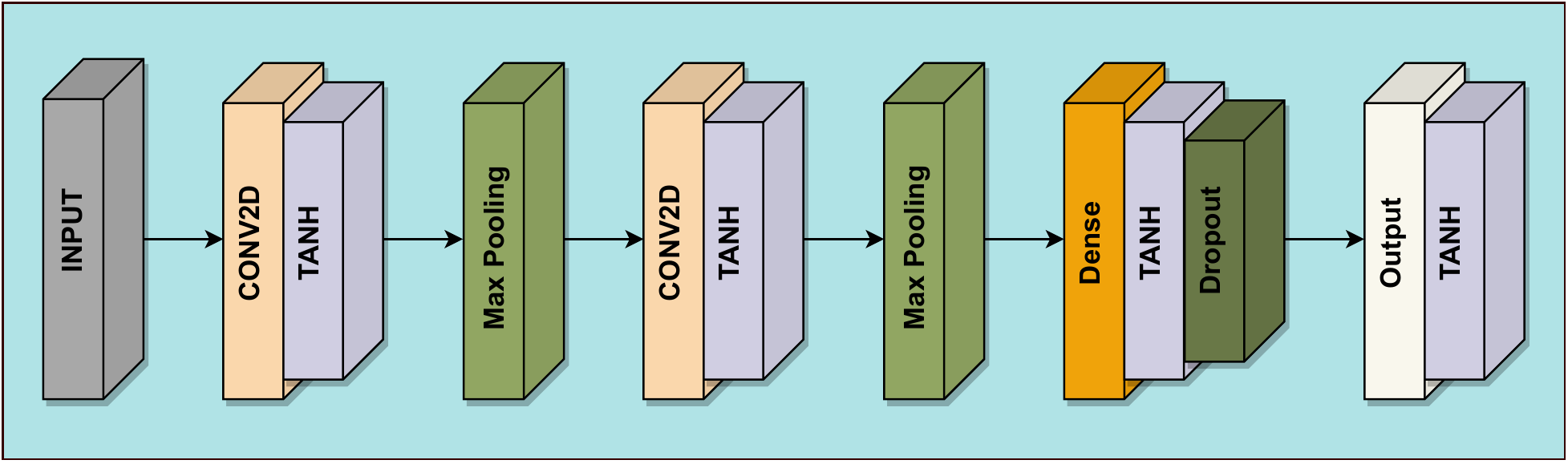}
    \caption{The architecture of the convolutional neural network for regression (CNN-REG).}
    \label{fig:CNN-REG}
\end{figure}

The first NN presented in~\cite{Mak2019} was the most straightforward NN based on convolutions. Hence, CNN-REG perfectly fits this DL model: a CNN for regression. The NN consisted of two convolutions and a dense layer mainly followed by the $tanh$ activation functions (as depicted in Figure~\ref{fig:CNN-REG}. The first convolutional layer had $48$ filters with kernel size  $3\times 15$, while the second convolution layer had 48 filters with kernel size $1\times9$. The dense layer consisted of $30$ neurons followed by the dropout layer with the dropout probability set to $0.25$. Furthermore, we have experimented with the values of the following hyperparameters to find the best possible performance:

\begin{itemize}
    \item max pooling kernel sizes: $1, 3, 5, 7, 10$,
    \item scaling factor $\kappa: 1, 5, 10$,
    \item Tested optimizers were $ADAM$ and $SGD$ both with learning rates $\alpha = 5\cdot10^{-3}, 5\cdot10^{-4}, 5\cdot10^{-5}$,
    \item Data augumentatation: \textit{row-wise} and \textit{batch-wise}.
\end{itemize}

Let us note that the NNs proposed by Mak et al. were enhanced by adding a scaling factor that increases the "width" of the NN. The width is enhanced by multiplying the number of filters in convolutional layers and the number of neurons in dense layers by scaling factor $\kappa$. For example, the initially proposed number of filters in the first convolutional layer was $48$, which after the scaling by $\kappa = 5$, becomes $240$. This way, we did not just compare our proposed AT-NN DL model with current state-of-the-art NNs; we have even tried to enhance the originally proposed models. Therefore we will present results for two versions of the CNN-REG NN topology: the original one (\textit{CNN-REG-ORG}) and the enhanced one (\textit{CNN-REG-ENH}).

The parameters for \textit{CNN-REG-ORG} are same as stated in~\cite{Mak2019} ($\kappa = 1$, max pooling kernel size equal to $3$) except we have used SGD with learning rate $\alpha = 0.005$ since the originally proposed learning rate $\alpha = 0.03$ would not converge on our dataset. As best dataset augmentation for this model was shown to be "Row-wise" data augmentation. For the \textit{CNN-REG-ENH} the best performing parameters were as follows: $\kappa = 10$, max pooling kernel size $5$, optimizer $ADAM$ with learning rate $\alpha = 0.0005$, and data augmentation \textit{row-wise}.

\subsubsection{Multivariate long short-term memory CNN (MLSTM-CNN): original and enhanced}
\label{sec2_3_2}
\vspace{10pt}

\begin{figure}
	\centering
	\includegraphics[clip, width=0.5\textwidth, trim=0cm 0cm 0cm 0cm]{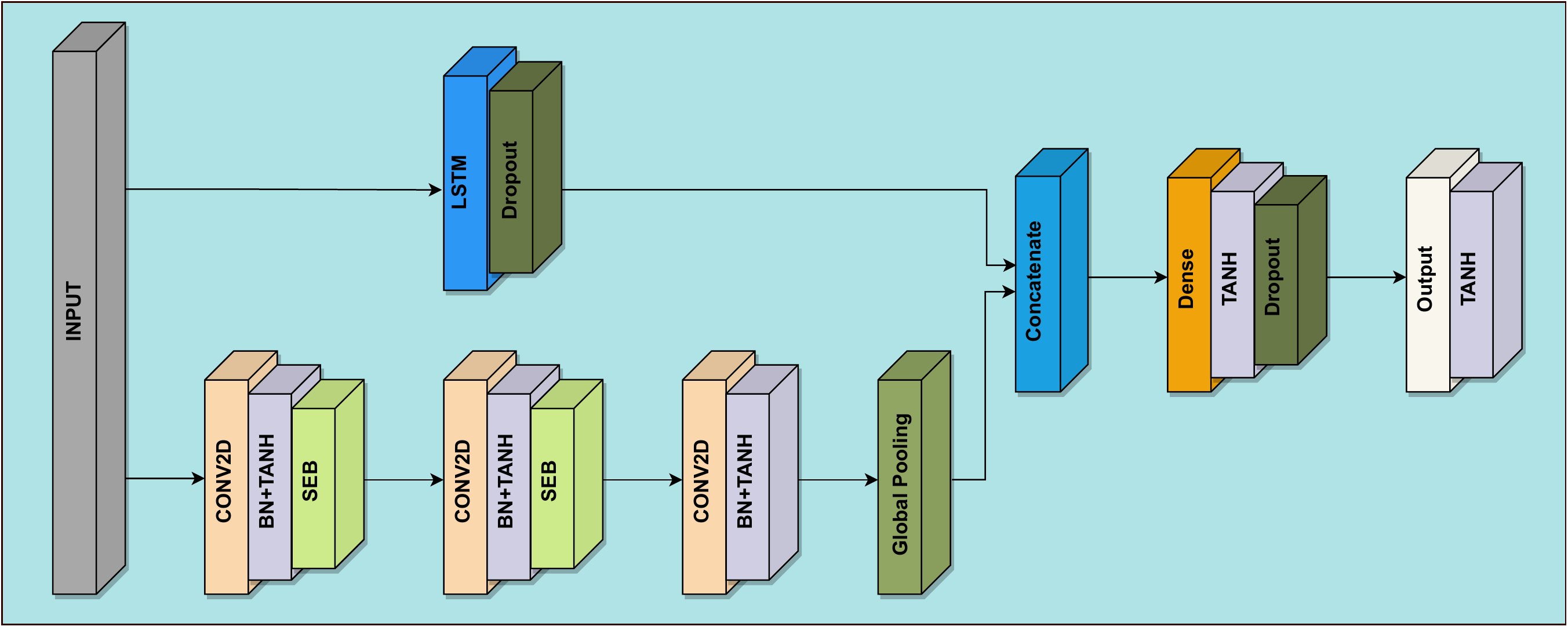}
    \caption{The architecture of the multivariate LSTM-CNN (MLSTM-CNN).}
    \label{fig:MLSTM-CNN}
\end{figure}

Multivariate long short-term memory (LSTM) CNN consists of two components, as depicted in Figure~\ref{fig:MLSTM-CNN}. The first component is based on convolutions (similar to the CNN-REG model) but enhanced with squeeze and excitation block (SEB)~\cite{Hu:2018:SEB}. SEB block introduces weights that increase inter-dependencies among the convolution layers feature maps while simultaneously adding minor computational costs. The second component is based on the LSTM block~\cite{Hochreiter:1997:LSTM}. The LSTM block introduces memory that is able to "remember" the dependencies between the signal sample in the observed moment and all samples before the observed moment. The multivariate long short-term memory convolutional neural network (MLSTM-CNN) is built the same way as proposed in \cite{Mak2019}: the first convolution layer had $16$ filters with kernel size $3\times11$, while the second and third convolution layer had $32$ filters with a kernel size of $1\times6$ and $1\times3$ respectively. The LSTM layer had depth $8$ (number of LSTM blocks stacked one on on top another) each with $32$ neurons (filters), and SEB block utilized "shrink" ratio of $r=16$. Finally, the dense layer consisted of $8$ neurons. This setup we have called \textit{MLSTM-CNN-ORG}. Next, by multiplying each number of neurons/filters by a factor $\kappa$, we created an enhanced version named \textit{MLSTIM-CNN-ENH}. 

The tested hyperparameters for MSLSTM-CNN are related to scaler value $\kappa$ chosen among the following options: $1$, $5$, and $10$. Other hyperparameters, such as data augmentation and optimizers (and their respective learning rates), were the same as in the case of CNN-REG.

After enhancing the originally proposed MLSTM-CNN, the best performing parameters for \textit{MSTM-CNN-ENH} were: data augmentation \textit{row-wise}, optimizer $ADAM$ with a learning rate set to $\alpha = 0.005$, and scaling factor $\kappa = 5$. On the other hand, the best performing parameters for \textit{MLSTM-CNN-ORG} were: data augmentation \textit{batch-wise}, and optimizer $SGD$ with learning rate $\alpha = 0.005$.

\subsubsection{Sliding puzzle network: original and enhanced}
\label{sec2_3_3}
\vspace{10pt}

\begin{figure}
	\centering
	\includegraphics[clip, width=0.5\textwidth, trim=0cm 0cm 0cm 0cm]{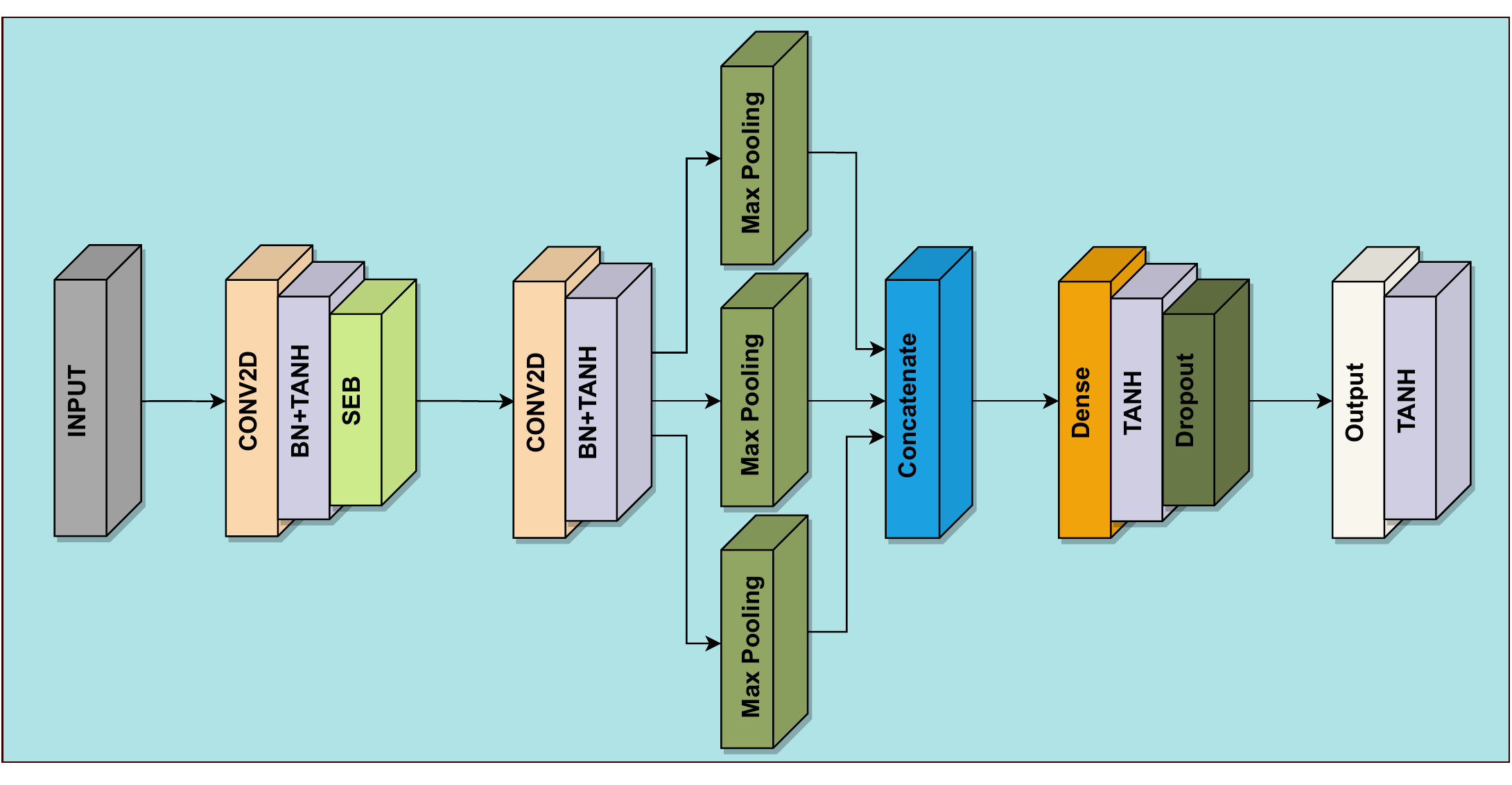}
    \caption{The architecture of the sliding puzzle neural network (SP-NN).}
    \label{fig:SlidingPuzzle}
\end{figure}

The sliding puzzle neural network (SP-NN) proposed in ~\cite{Mak2019} has a pretty neat motivation. This model aims to shuffle random patches of the input data in a manner that resemble a sliding puzzle. In a sliding puzzle, although the image is shuffled, the content of the image can still be seen. This idea led to the construction of the NN, which topology can be seen in Figure~\ref{fig:SlidingPuzzle}. The SP-NN is designed to remove the positional dependency between samples completely. To achieve this, the average, max, and min pooling are performed for each feature map of the second convolution layer. As the original authors state: the result of encoding is purely based on initial convolutions.

We have built two versions of SP-NN: the original one that resembles the model proposed by its authors (\textit{SP-NN-ORG}) and an enhanced one (\textit{SP-NN-ENH}) where we increased the width of \textit{SP-NN-ORG} by a factor $\kappa$. Therefore \textit{SP-NN-ORG} had the following hyperparameters: the first convolution layer had $64$ filters with kernel size $1\times25$, the second convolution layer had $128$ filters with kernel size $3\times1$, dense layer consisted of $30$ neurons, SEB block had shrink ratio of $16$. Since the authors did not mention the kernel sizes of the max, average, and min pooling layers, we have tested the following kernel sizes: $1\times1$, $3\times3$, $5\times5$, $7\times7$, and $10\times10$ for both versions. Furthermore, we have trained SP-NN without any data augmentation, as well as with "row-wise" and "batch-wise" augmentation. The tested optimizers and respective learning rates were the same as for CNN-REG and MLSTM-CNN.

For the \textit{SP-NN-ENH} we have scaled a number of filters and neurons by multiplying them with factor $\kappa = 10$. Optimizer for \textit{SP-NN-ENH} was adam with $\alpha = 0.00005$. The optimizer for the original version was SGD with $\alpha = 0.005$. 

We believe that we have enhanced all NNs in a manner that does not violets the original author's reasons and motivation behind NN design. Also, wherever we had doubts about parameters' values (when these were not specified in the original work), we tested different parameter values and took the best-performing one. 

\subsubsection{Attention-based neural network (AT-NN)}
\label{sec2_3_4}
\label{AT-NN}
\vspace{10pt}

Many state-of-the-art algorithms in computer vision, natural language processing (NLP), or even augmented data generation rely on attention modules~\cite{Attention:NLP,pmlrv97zhang19d:SelfAttention-GAN, Woo_2018_ECCV:CBAM}. The attention module, including the self-attention module, has been introduced in paper~\cite{NIPS2017:SelfAttention}, which claims that "Attention is all you need." The intuition behind the attention module comes from NLP, where the urge to capture relations between words far away from each other in a sentence/text was necessary. Namely, one word can contain crucial information that, in combination with another word (for instance, a subject in the sentence), would completely change the sentence's meaning. However, this information can be spread along the whole text, so the NNs that are based on memory cells (LSTM~\cite{YU:2019:LSTM}) or windows (CNN) would be incapable of capturing it. 

We believe that the time-series signals describing the ship's motion have also scattered information that needs to be grouped. The proposed model AT-NN can seize the interaction and correlation between different parts of the input signals by utilizing the self-attention module. Thus, the model can create new informative features essential for sea state estimation, which simple convolution operations can not obtain.

\begin{figure*}
	\centering
	\includegraphics[clip, width=\textwidth, trim=0cm 0cm 0cm 0cm]{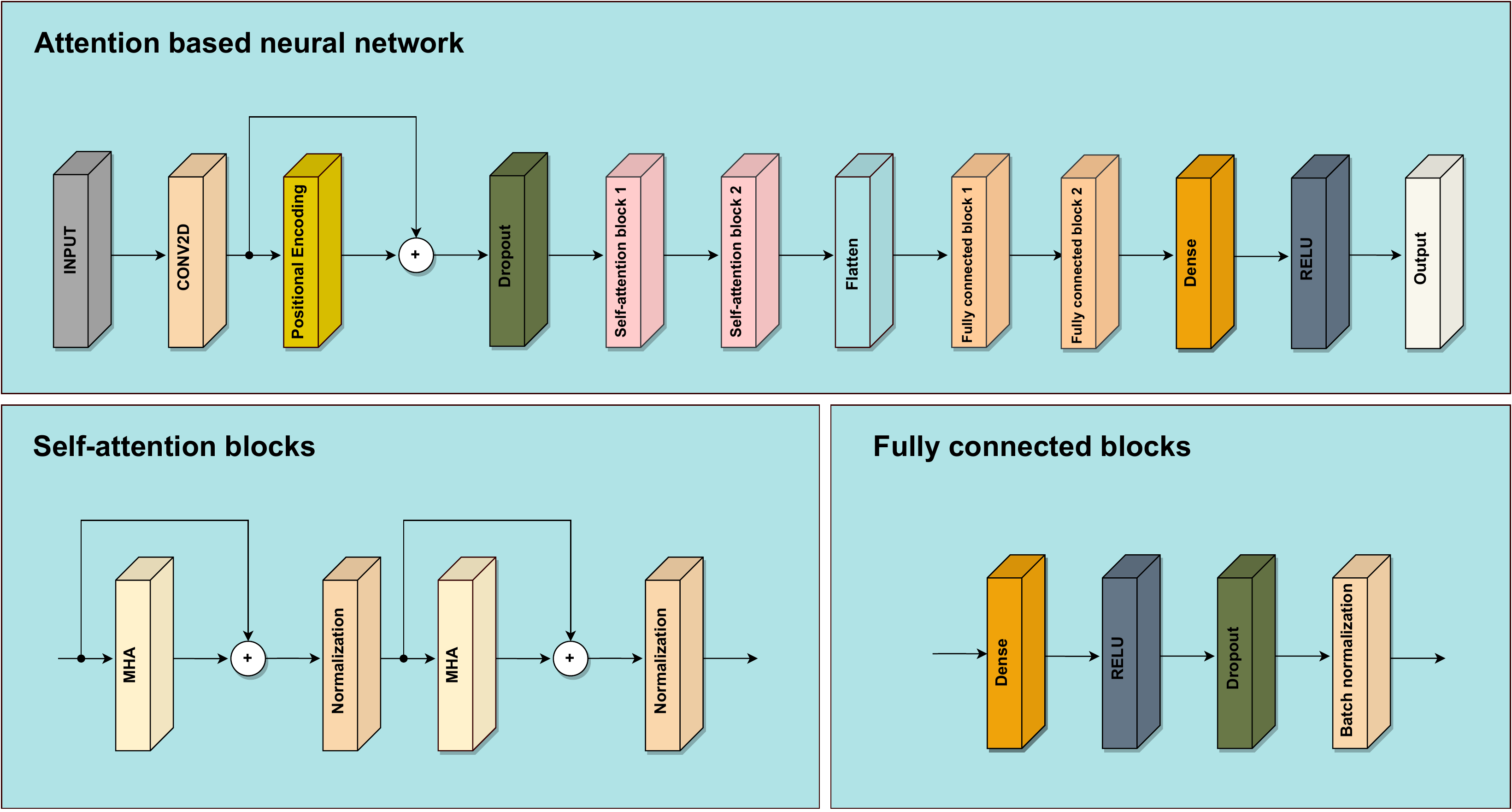}
    \caption{The architecture of the attention based neural network (AT-NN).}
    \label{fig:AT-NN}
\end{figure*}

The architecture of the proposed AT-NN is depicted in Figure~\ref{fig:AT-NN}. At the beginning of the NN, the heave, pitch, and yaw signals are merged into one embedded signal by performing 2D convolution with kernel size $3 \times 125$. Since we utilized $128$ kernels in this operation, NN operates on $128$ embedded signals instead of three original ones. Besides merging three signals into one, the convolution operation also resulted in $1,377$ overlapping window positions over the original signals. Each window captured $125$ samples of the input signals. These overlapping window positions are equivalent to the "tokens" in the NLP models. Established practice in NLP recommends that each token have its position in sentence/signal encoded because the inter-token positions may be relevant information~\cite{NIPS:2015:transforemer}. We have followed this practice and utilized the sin and cosine encoding as proposed in~\cite{NIPS2017:SelfAttention}.

After merging and tokenizing the input signals, the proposed AT-NN 
utilizes two self-attention blocks. As depicted in Figure~\ref{fig:AT-NN}, each self-attention block consists of two multi-head attention (MHA) layers followed by a normalization layer. The attention mechanism is given by the equation ~\ref{eq:attention}:

\begin{equation}\label{eq:attention}
attention(Q,K,V) = softmax(\frac{QK^T}{\sqrt{d_k}})V
\end{equation}

where $Q$, $K$, and $V$ represent query, key, and value matrices, respectively, with their weights included ($W_q, W_k, W_v$). In our case, we used self-attention where $Q$, $K$, and $V$ were all equal to the layer's input. Generally speaking, the self-attention mechanism seeks relations between the tokens in the same input signal, not between the observed input signal and another different signal. Parameter $d_k$ is a scaling factor equal to the length of the input embedding (in our case, $1,377$). 

On the other hand, the MHA is a mechanism that applies the self-attention mechanism several times in parallel, learning different relations between tokens simultaneously. The MHA is given by the equation~\ref{eq:mha}:

\begin{equation}\label{eq:mha}
\begin{aligned}
        multihead(Q,K,V) &= concat(head_1, ... , head_i) W^l\\
        head_i & = attention(Q, K, V ) 
\end{aligned}
\end{equation}

Each of $i=2$ heads uses its own weights $W_{q}^i, W_{k}^i,$ and $W_{v}^i$, which consequently means that each head learns different relations between tokens. Finally, heads' outputs are concatenated into one embedding. Embedding is scaled to match the original input "embedding" size by performing a simple linear projection utilizing weights $W^l$. Furthermore, the output of MHA is added to its input. After the addition is performed, the final summed output of the MHA is normalized. Normalization is done by subtracting the embedding mean value from the embedding values and dividing it by their standard deviation~\cite{Garbin:2020:normalisation}. The described mechanism is repeated two times in one self-attention block, as depicted in Figure~\ref{fig:AT-NN}. AT-NN model was built from two consecutive self-attention blocks. 

The final embedding calculated by the second MHA ($batch_{size} \times 128 \times 1,377$) is flattened into $batch_{size}$ number of 1D vectors with a length of $176,256$. This is important because the next part of the proposed AT-NN consists of fully-connected (dense) layers. As depicted in Figure~\ref{fig:AT-NN}, the AT-NN consists of three dense layers. The first dense layer has $128$ neurons and is followed by the RELU activation function, dropout layer, and batch normalization layer~\cite{Abien:2018:RELU, Santurkar:2018:BatchNorm}. The second dense layer had $64$ neurons, followed by the RELU activation function, dropout layer, and batch normalization layer. The last dense layer had only three neurons. Each of the three neurons represents outputs of the AT-NN: $H_s$, $T_z$, and $\beta$, respectively. The final output is one more time activated by the RELU activation function to avoid small negative number predictions. It is necessary to mention that all dropout layers (everywhere in AT-NN) had dropout probability set to $0.1$ ~\cite{Baldi:2013:dropout}. 

Speaking of dropout layers, besides regularization purposes, they play a key role in the uncertainty estimation of the NN's decision. Uncertainty estimation raises the trustworthiness of models' predictions. The dropout method's uncertainty estimation is based on Monte-Carlo sampling of different variations of the original NN during the prediction. As proved in ~\cite{Gal:2016:MCDO}, the sampled NNs can be treated as approximate Bayesian inference in deep Gaussian processes. To the best of our knowledge, our model is the first to introduce uncertainty estimation to the DL models for sea state estimation.

All parameter selection is made after a computationally heavy best-parameter combination search. That being said, we have considered the following values as the candidates for chosen hyperparameters: 
\begin{itemize}
    \item number of MHA blocks: $2,4,8$,
    \item number of heads in each MHA block: $2, 4, 8$,
    \item token size: $125, 250, 500$,
    \item number of embeddings: $32, 64, 128, 256$,
    \item number of neurons in fully connected layers consecutively: $(128, 64, 3)$, $(256, 128, 3)$, $(512, 256, 3)$, and $(1,024, 512, 3)$,
    \item Tested optimizers were $ADAM$ and $SGD$ both with learning rates $\alpha = 5\cdot10^{-3}, 5\cdot10^{-4}, 5\cdot10^{-5}$~\cite{Kingma:2014:Adam,Bottou:2012:SGD},
    \item Data augumentatation: \textit{row-wise} and \textit{batch-wise}.
\end{itemize}
The model was trained with $batch_{size}$ set to $32$ for $250$ epochs with early stopping if there was no change in loss function on the validation data set for $10$ epochs. The best model consisted of 2 MHA blocks with $2$ heads each, a token size of $125$, and the number of embeddings set to $128$. The number of neurons in dense layers was $(128, 64, 3)$ respectively, while the optimizer was ADAM with a learning rate set to $5\times10^{-4}$.


\subsection{Hyperparamter overview and training setup}
\label{sec2_4}
\vspace{10pt}

The overview of the best-performing hyperparameters for each model is given in Table~\ref{table:hyperparameters}. Since we trained a large number of models, the best performance (top 25), conducted with different hyperparameters, was achieved by those that used some kind of data augmentation. However, there is no evident distinction between the two types of augmentation (batch-wise and row-wise) in terms of the produced results.

The loss function was mean squared error (MSE)~\cite{James:1992:MSE}. The training process mentioned in section ~\ref{AT-NN} was the same for all models: batch size was always set to $32$, with early stopping after there was no improvement on validation loss of $10$ epochs. The constraint that the models are training for $250$ epochs was never fulfilled because all models converged after the $100$ to $150$ epoch.

\begin{table*}[!ht]
    \centering
    \caption{An overview of the best performing hyperparameters
for each model.}
    \resizebox{\textwidth}{!}{
    \begin{tabular}{|c|c|c|c|c|c|c|c|c|c|c|}
    \hline
       \textbf{Model} & \textbf{DATA\_AUG} & \textbf{OPTIMIZER} & \textbf{LR} & \textbf{SCALE} & \textbf{MAXPOOL} & \textbf{BLOCKS} & \textbf{NUM HEADS} & \textbf{EMB SIZE} & \textbf{NUM EMB} & \textbf{HEADS NEURONS} \\ \hline
        CNN-REG-ORG & Row-wise & SGD & 0,005 & 3 & 1 & / & / & / & / & / \\ \hline
        MSTM-CNN-ORG & Batch-wise & SGD & 0,005 & 1 & / & / & / & / & / & / \\ \hline
        SP-NN-ORG & Batch-wise & SGD & 0,005 & 5 & 1 & / & / & / & / & / \\ \hline
        CNN-REG-ENH & Row-wise & ADAM & 0,0005 & 10 & 5 & / & / & / & / & / \\ \hline
        MLSTM-CNN-ENH & Row-wise & ADAM & 0,005 & 5 & / & / & / & / & / & / \\ \hline
        SP-NN-ENH & Batch-wise & ADAM & 0,00005 & 10 & 5 & / & / & / & / & / \\ \hline
        AT-NN & Batch-wise & ADAM & 0,0005 & / & / & 2 & 2 & 125 & 128 & 128,64,3 \\ \hline 
    \end{tabular}
    }
    \label{table:hyperparameters}
\end{table*}

Next, we provide the achieved results for all tested models and elaborate on their efficiency, advantages, and limitations. 

\section{Results and discussion}
\label{sec3}
\vspace{10pt}

\subsection{NN models performance}
\vspace{10pt}
\label{sec3_1}

Evaluation of model performances is done separately for each sea state parameter: $H_s$, $T_z$, and $\beta$. We have calculated two metrics: MSE and mean absolute error (MAE)~\cite{Chai:2014:MSEMAE} on both validation and test sets. The motivation behind utilizing these metrics as evaluation is in their nature. Namely, when training NNs, it is necessary to focus the models on the most troublesome samples. This behaviour can be achieved through MSE because it makes small errors (correct predictions) even smaller. At the same time, it increases the influence of errors in samples that the model did not predict correctly (or was the furthest from the correct prediction). On the other hand, we have also utilized MAE, which may be considered a reliable and realistic metric for the real-life scenario of sea state estimation.

\begin{figure}
\centering
\begin{subfigure}[b]{0.49\textwidth}
   \includegraphics[width=1\linewidth]{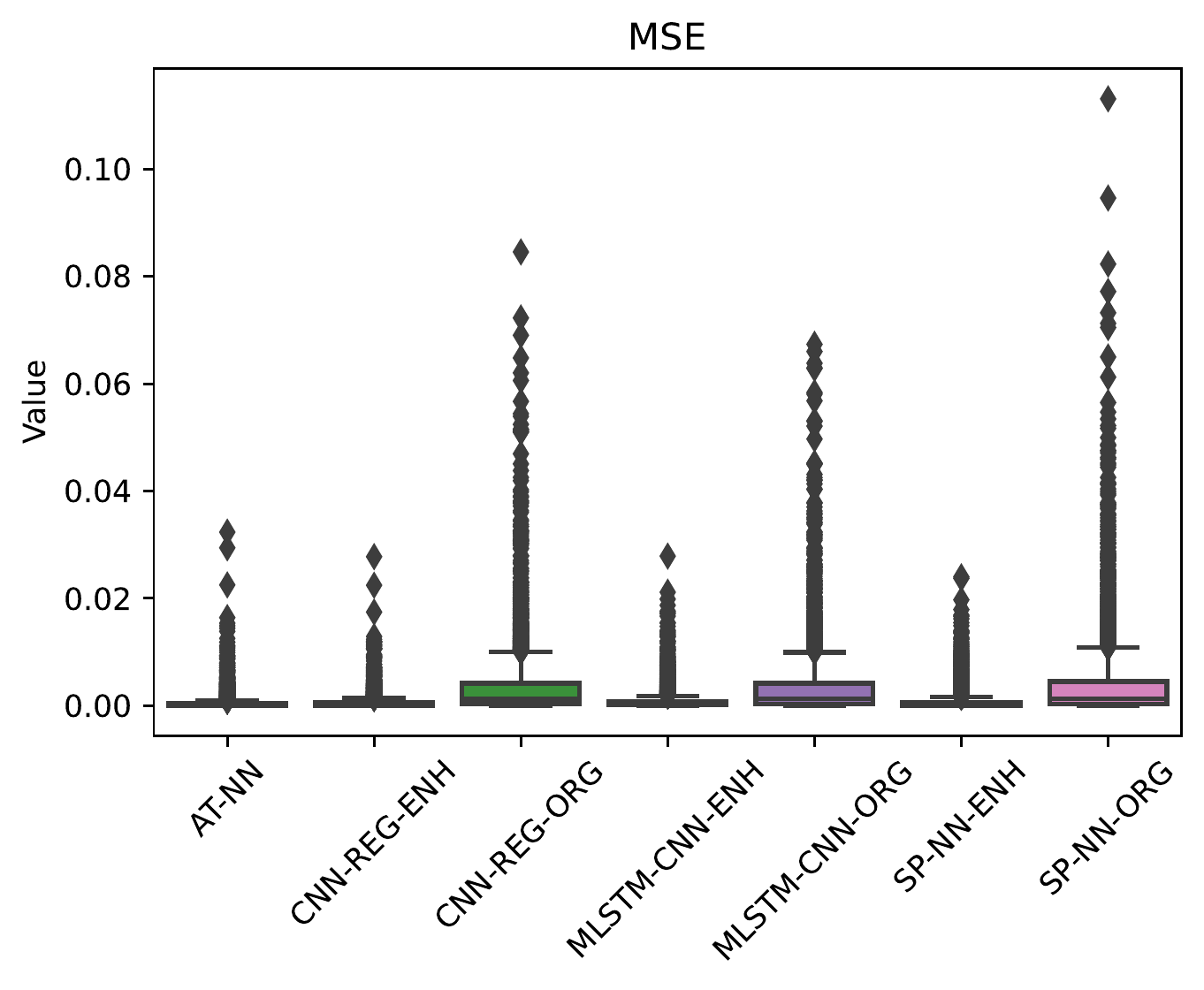}
   \caption{}
   \label{fig:boxplot-MSE} 
\end{subfigure}

\begin{subfigure}[b]{0.49\textwidth}
   \includegraphics[width=1\linewidth]{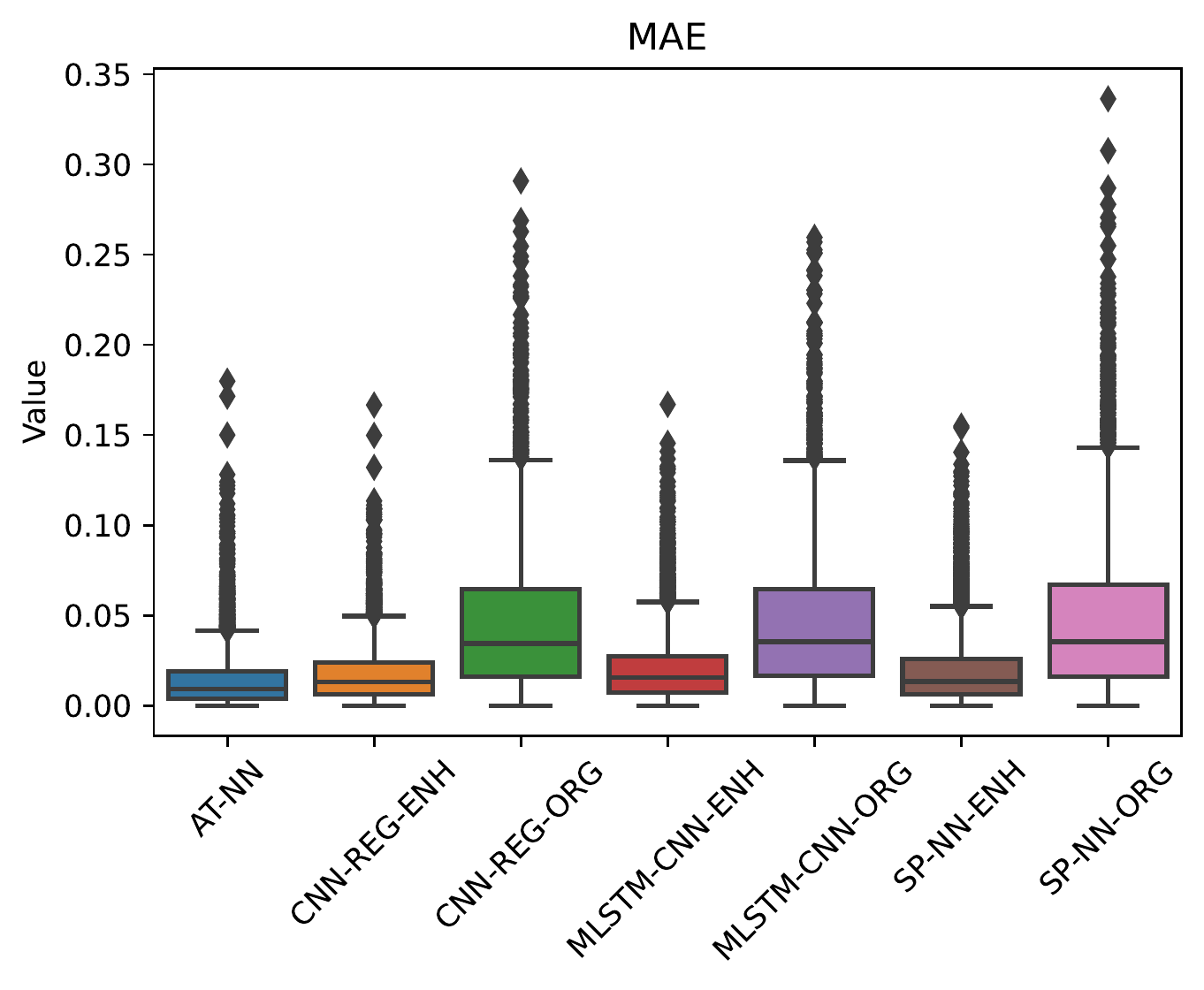}
   \caption{}
   \label{fig:boxplot-MAE}
\end{subfigure}
\caption[Two measures]{Box-plot diagrams on the test set with two measures: a) MSE and b) MAE.}
\label{fig:boxplot}
\end{figure}

Tables~\ref{table:ValidationResults} and~\ref{table:TestResults} present model results for each estimated parameter on validation and test set respectively. As can be seen, columns named MSE and MAE represent the average MSE and average MAE of the three estimated sea parameters (showing the general score of the models over all three estimated sea state parameters).

By inspecting the difference of respective scores between the validation and test set (Tables~\ref{table:ValidationResults} and~\ref{table:TestResults}), it can be noticed that the models' performances have small deviations. This proves that models were appropriately trained and that early stopping indeed stops the models from over-fitting. Further investigation of the results shows a noticeable difference in performance between the enhanced group of models (including AT-NN) and the original models proposed in \cite{Mak2019}. This reflects all estimated sea state parameters and general scores. The difference between original and enhanced models (including the proposed model AT-NN) is quite large: by a factor $~5\times$. For instance MSE \textit{CNN-REG-ORG} is $0.0035$, while the MSE of \textit{CNN-REG-ENH} is $0.00059$. The same applies to all other models on both validation and test sets. To support our claims and make results easier to follow, in Figure~\ref{fig:boxplot} we present box-plot diagrams of MSE and MAE on the test set. From the provided box plots, it is evident that we have successfully enhanced the original models.

\begin{figure*}
	\centering
	\includegraphics[clip, width=\textwidth, trim=4cm 6cm 4cm 6cm]{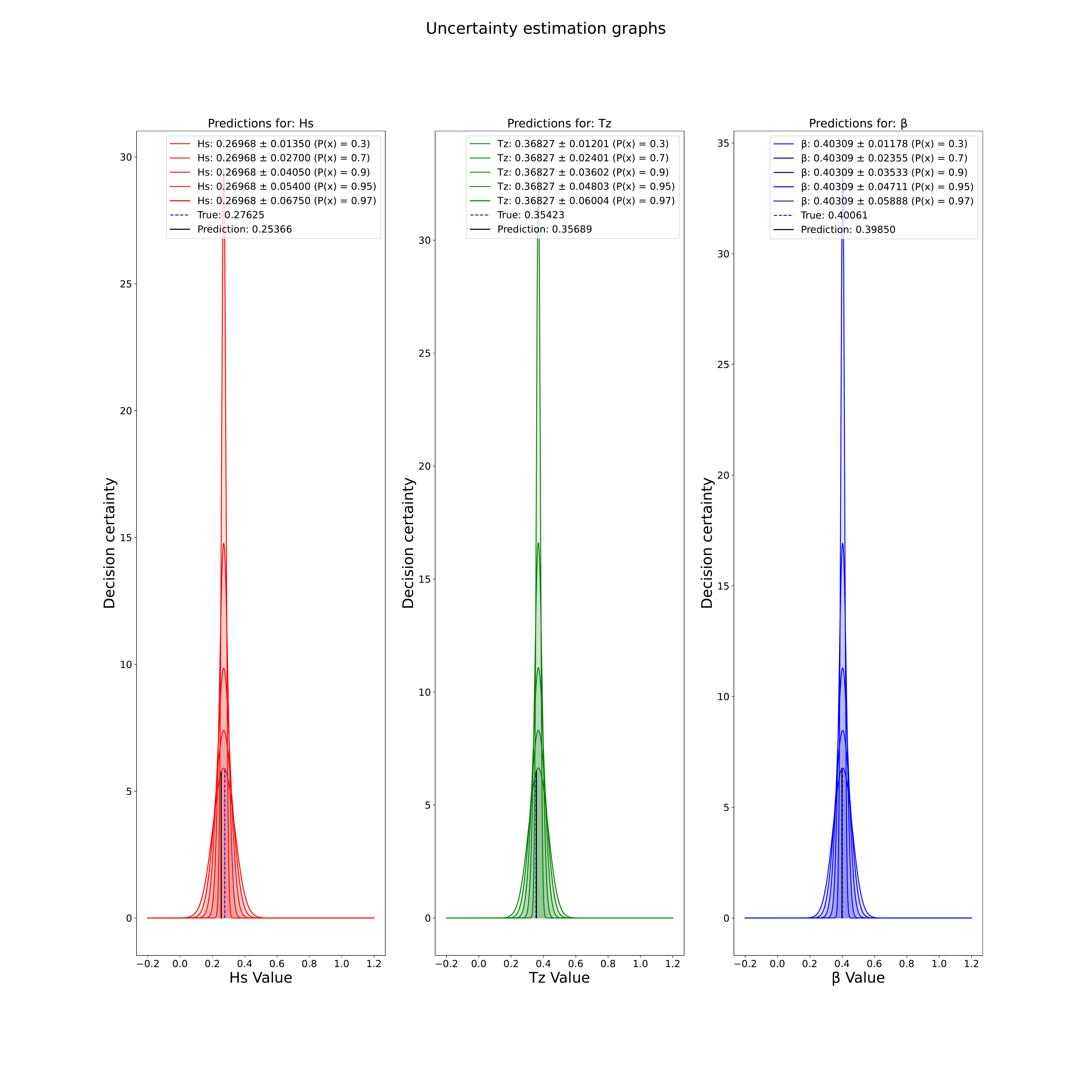}
    \caption{Uncertainty estimation of NN outputs $H_s$, $T_z$ and $\beta$ based od Monte-Carlo dropout.}
    \label{fig:System2}
\end{figure*}
\begin{table*}[!ht]
    \centering
     \caption{Models results of estimated parameters on validation dataset ($\times 10^-3$).}\label{table:ValidationResults}
    \resizebox{\textwidth}{!}{
    \begin{tabular}{|c|c|c|c|c|c|c|c|c|}
        \hline
    \multirow{2}{*}{\textbf{Model}} & \multicolumn{3}{c|}{\textbf{MSE}} &\multicolumn{3}{c|}{\textbf{MAE}} & \multirow{2}{*}{\textbf{MSE}} & \multirow{2}{*}{\textbf{MAE}} \\ \cline{2-7} & $H_s$ & $T_z$ & $\beta$ &$H_s$ & $T_z$ & $\beta$ & & \\ 
    \hline
    
     CNN-REG-ORG & 5,28544 & 3,889443 & 1,252465 & 57,042372 & 50,812191 & 26,702811 & 3,475783 & 37,110115 \\ \hline
        MLSTM-CNN-ORG & 4,941625 & 4,194312 & 1,113016 & 56,162481 & 53,026415 & 25,387024 & 3,416318 & 37,535071 \\ \hline
        SP-NN-ORG & 5,225949 & 4,765217 & 1,413309 & 57,096877 & 54,959986 & 26,934951 & 3,801492 & 38,619452 \\ \hline
        CNN-REG-ENH & 0,93894 & 0,478358 & 0,210208 & 23,320904 & 16,435646 & 11,417285 & 0,542502 & 13,433017 \\ \hline
        MLSTM-CNN-ENH & 1,259706 & 0,604518 & 0,29818 & 26,476073 & 19,144859 & 13,941651 & 0,720801 & 15,447244 \\ \hline
        SP-NN-ENH & 1,289881 & 0,75156 & 0,169424 & 26,581532 & 19,68622 & 9,921086 & 0,736955 & 15,668236 \\ \hline
        AT-NN & \textbf{0,802887} & \textbf{0,372965} & \textbf{0,115358} & \textbf{20,823295} & \textbf{13,835668} & \textbf{7,599803} & \textbf{0,430403} & \textbf{11,696455} \\ \hline

    \end{tabular}
    }
\end{table*}

\begin{table*}[!ht]
\label{tbl1}
    \centering
     \caption{Models results of estimated parameters on test dataset ($\times10^-3$).}\label{table:TestResults}
    \resizebox{\textwidth}{!}{
    \begin{tabular}{|c|c|c|c|c|c|c|c|c|}
        \hline
    \multirow{2}{*}{\textbf{Model}} & \multicolumn{3}{c|}{\textbf{MSE}} &\multicolumn{3}{c|}{\textbf{MAE}} & \multirow{2}{*}{\textbf{MSE}} & \multirow{2}{*}{\textbf{MAE}} \\ \cline{2-7} & $H_s$ & $T_z$ & $\beta$ &$H_s$ & $T_z$ & $\beta$ & & \\ 
    \hline
    CNN-REG-ORG & 5,039482 & 3,939628 & 1,619829 & 55,412124 & 50,962495 & 28,444887 & 3,53298 & 36,635866 \\ \hline
        MLSTM-CNN-ORG & 4,886629 & 4,118302 & 1,555991 & 55,245936 & 52,170407 & 28,296627 & 3,520307 & 36,978883 \\ \hline
        SP-NN-ORG & 5,262113 & 4,662741 & 1,902819 & 58,077689 & 53,857223 & 28,712736 & 3,942557 & 38,625823 \\ \hline
         CNN-REG-ENH & 0,981791 & 0,561467 & 0,231101 & 24,165625 & 17,323201 & 11,694066 & 0,591453 & 14,02676 \\ \hline
        MLSTM-CNN-ENH & 1,345254 & 0,691574 & 0,290011 & 26,954267 & 20,597859 & 13,599609 & 0,775613 & 16,109246 \\ \hline
        SP-NN-ENH & 1,25686 & 0,861854 & 0,216848 & 26,767925 & 21,339719 & 10,730237 & 0,77852 & 16,295388 \\ \hline
        AT-NN & \textbf{0,851608} & \textbf{0,404057} & \textbf{0,114521} & \textbf{21,122448} & \textbf{13,955757} & \textbf{7,485123} & \textbf{0,456728} & \textbf{11,844978} \\ \hline

    \end{tabular}
    }
\end{table*}
Since the dataset in the paper ~\cite{Mak2019} was not publicly available, we could not apply our method to it, and we had to re-implemented all of the models. By comparing their loss values results on the validation set (MSE = $~0.05$), we can conclude that produced results using the original models are improved. In our study, instead of 6-DOF motions, only 3-DOF was used as input parameters. Accordingly, we successfully demonstrated that original models achieve the same or better results with reduced input data. This is also, besides estimation accuracy, one of the advantages of here proposed approach.

Next, we point out the fact that our proposed model achieved the best results among all tested models. In both tables \ref{table:ValidationResults} and ~\ref{table:TestResults}) the best results on each estimated parameter are bolded. Therefore, the best overall MSE on the test set was MSE(AT-NN) = $0.00046$ for the proposed model with its MAE(AT-NN) = $0.012$. The runner-up model was \textit{CNN-REG-ENH} model with MSE = $0.00059$ and MAE = $0.014$. Since the error was already small ($<10^{-3}$) achieved by existing models, the improvement made by the proposed model AT-NN was impactful. In other words, the AT-NN improved the MSE by $22.778\%$ and MAE by $15.554\%$ over the best-performing enhanced model.

According to the results in Table \ref{table:TestResults}, one can see that the proposed AT-NN model achieves up to 94\% improvement of MSE, and up to 70\% improvement of MAE, compared to the original models (denoted by CNN-REG-ORG, MLSTM-CNN-ORG, and SP-NN-ORG). Also, the enhanced models performed better than the above-mentioned original models, so a valid comparison is a comparison between the proposed model and these enhanced models. The enhanced SP-NN model provided the best estimation of the angle $\beta$ if we observe MSE and MAE for this parameter. However, the AT-NN model can achieve even more pronounced improvement. Namely, the MSE value is improved by 47\% and MAE by 30\% compared to SP-NN-ENH. This improvement is also the largest in terms of estimation of angle $\beta$,  achieved with the AT-NN model.

When analyzing the estimation results for $H_s$ and $T_z$, CNN-REG-ENH distinguished out among the enhanced models with the best results. Again, AT-NN improves the results of the best-enhanced model. The proposed model improve MSE 13\% and MAE 13\% for $H_s$, MSE 28\% and MAE 19\% for $T_z$ parameter in comparison to the best performing enhanced models. In comparison to the original models, the improvement is even more significant in all the observed parameters. Observing the performance of the AT-NN model, the estimation of the period $T_z$ and the angle $\beta$ achieves the most favorable advancements compared to the estimation of the parameter $H_s$.

To increase the trustworthiness of the proposed model, we have enabled uncertainty estimation as an additional feature.

\subsection{Uncertainty estimation}
\label{sec3_2}
\vspace{10pt}

As mentioned in subsection~\ref{AT-NN}, adding dropout layers to the "head" part of the AT-NN model enabled the utilization of Monte-Carlo sampling. By utilizing Monte-Carlo dropout sampling, we simulated the Gaussian process and, therefore, obtained model prediction uncertainty for each estimated sea state parameters~\cite{Gal:2016:MCDO}. The process is quite simple in bolt and nuts: for the same input time-series signals, we run predictions of the AT-NN $256$ times, on every prediction dropping random neurons. By dropping random neurons, we obtain a new version of the original NN that gives predictions slightly different from other "sampled" networks. By calculating the mean value $\mu$ standard deviation $\sigma$ of all predictions, we can calculate a Gaussian curve that estimates the uncertainty of the model decision. Accompanying uncertainty estimation, we also have the actual prediction of the model generated without dropping any of the neurons.

To increase explainability and provide additional information, we have extended the output of Monte-Carlo dropout sampling by involving statistical measurement. Although quite simple in its nature, statistical measurement involved counting how many correct measurements of the test set fell into Gaussian curves estimated by Monte-Carlo dropout sampling with different values of $\sigma$ ($\mu$ was fixated). The different values of $\sigma$ were calculated by multiplying sigma with $n \in [1,2,3,4,5]$ resulting in wider Gaussian curves. Thus, following one's expectation, wider Gaussian curves would also encompass a bigger number of correct measurements of the sea state. Therefore for $n = 1$, the Gaussian curve with $\sigma_\text{scaled} = 1 \times \sigma$ contained $387$ correct predictions while $896$ predictions/samples were outside the range. This leads us to $~30\%$ certainty in total ($387 / 1283$) for this quite narrow, Gaussian curve. On the other hand, a much wider Gaussian curve with $\sigma_\text{scaled} = 5 \times \sigma$ contained $1246$ predictions from $1283$ samples in test set which lead us to the certainty of $~97\%$. The validation dataset had similar performance related to $\sigma$ values and the number of contained samples, so the integrity of the estimation was not impaired. The graphical example of the final AT-NN output is given in Figure~\ref{fig:System2}. It can be seen that the blue line represents the true value for the given input signals, and the black line represents the prediction of the AT-NN without dropping any neurons.
In contrast, Gaussian curves represent different "intervals"/"areas" of probability that correct measurements lie within. We believe that this novel process of estimating and interpreting the NN's output raises the trustworthiness of the proposed model. Also, besides having only one number, as a result, the user of the proposed model, if having doubts in model estimation, now has confidence levels provided by different Gaussian curves. We believe that uncertainty estimation tailored in the previously described way can be crucial information in various practical scenarios. 

\subsection{Summary of contributions and novelty}
\label{sec3_3}
\vspace{10pt}

Let us conclude our study by clearly pointing out the achievements and novelty introduced in this manuscript, summarized as follows:
\begin{itemize}
    \item We have retrained the existing state-of-the-art models proposed in~\cite{Mak2019} and enhanced them by increasing their width. The enhanced models performed better than the originally proposed models.
    \item We have proven that it is possible to obtain similar or better results by only 3-DOF ship motion responses instead of using 6-DOF ship motion responses (sea state parameters' estimation improved by 23\% in terms of MSE and by 16\% in terms of MAE).
    \item As guidelines for model training, we have dug into data augmentation methods based on shuffling input signal slices. The investigation concludes that all tested models achieve better results with shuffled input signals' slices.
    \item We have proposed a novel AT-NN model that achieved the best results among all tested models for each estimated sea state parameter. To the best of our knowledge, this is the first time the AT-NN to be implemented for sea state estimation. 
    \item Furthermore, in this manuscript, we have proposed a novel way of interpreting the uncertainty estimation of NN outputs based on the Monte-Carlo dropout method. The way we interpret the uncertainty of the model's predictions would raise the model's trustworthiness even if his initial prediction achieved a negligible error rate. Also, this is the first time for the uncertainty estimation of the DL models in the sea state estimation, according to the extensive literature review. 
\end{itemize}

\section{Conclusion}
\label{sec4}
\vspace{10pt}
The paper presents a novel attention-based DL model for estimating important sea state characteristics. By evaluating the performance for each sea state parameter, we confirm that the AT-NN model is fine-tuned and suitable for estimating key sea state parameters. The improvement made by the proposed AT-NN model proved significant as MSE was reduced by about 23\% and MAE by about 16\%, compared to the best versions of the original, formerly proposed state-of-the-art models. Despite using 3-DOF input motion data, the proposed model achieved the best results among all tested, original and enhanced models for each estimated sea state parameter. In practical applications, entire wave directional spectrum estimation is of great importance. Therefore, future work directions may include information about wave directionality in numerical simulations that can extend the application of the DL model to short-crested waves and combined wind and swell sea spectrums.

\section*{Declaration of competing interest}
The authors declare no conflict of interest.

\printcredits

\section*{Funding}

This work was fully supported by the Croatian Science Foundation under the project IP-2018-01-3739, EU Horizon 2020 project "INNO2MARE" under the number 101087348, IRI2 project ”ABsistemDCiCloud” (KK.01.2.1.02.0179), University of Rijeka projects uniri-tehnic-18-17 and uniri-tehnic-18-15 and the European COST project CA17137.

\bibliographystyle{cas-model2-names}

\bibliography{references}



\end{document}